\documentclass[letterpaper, 10 pt, conference]{ieeeconf}  
\usepackage{graphicx}
\usepackage{amsmath,amssymb,amsfonts}
\usepackage{booktabs}
\usepackage{textcomp}
\usepackage{xcolor}
\usepackage{cite}
\usepackage{multirow}
\usepackage{subcaption}
\usepackage{url}
\usepackage{cuted} 

\usepackage[ruled,vlined,linesnumbered]{algorithm2e}

\makeatletter
\let\NAT@parse\undefined
\makeatother

\usepackage{hyperref}
\definecolor{lightblue}{rgb}{0.68, 0.85, 0.9}
\definecolor{darkblue}{rgb}{0.48, 0.65, 0.7}
\definecolor{lightgray}{gray}{0.9}
\definecolor{citecolor}{HTML}{0071BC}
\hypersetup{colorlinks,linkcolor={red},citecolor={citecolor}}  

\IEEEoverridecommandlockouts

\title{\LARGE \bf
Frontier-as-Token: Structured Spatial Reasoning \\ for Zero-shot Object Navigation
}

\author{Mulang Shi$^{1}$ and Mingyu Ding$^{1}$%
\thanks{$^{1}$Mulang Shi and Mingyu Ding are with the Department of Computer Science,
University of North Carolina at Chapel Hill.
{\tt\small \{mlshi, md\}@cs.unc.edu}}%
}
\newcommand{\methodname}[1]{FaT}

\begin{document}

\maketitle
\thispagestyle{empty}
\pagestyle{empty}

\begin{abstract}
Zero-shot object navigation requires an embodied agent to explore unseen environments and reliably stop upon locating a target category.
Recent approaches integrate vision-language models (VLMs) with frontier-based exploration, but remain limited by three structural issues: i) collapsing spatial frontier candidates into scalar scores, ii) relying on brittle single-frame stopping decisions, and iii) employing heuristic recovery strategies that fail to prevent repeated false positives.
We propose Frontier-as-Token (\methodname{}), a unified decision framework that reformulates frontier exploration as structured token-level reasoning with probabilistic commitment. Instead of treating frontiers as transient geometric points, \methodname{} assigns persistent identities to frontier tokens and constructs a structured multimodal representation that enables comparative VLM reasoning over spatially grounded candidates.
For termination, \methodname{} replaces threshold-based stopping with Bayesian evidence accumulation, integrating detection confidence, geometric consistency, and VLM verification into a sequential belief update. To mitigate cyclic false detections, we further introduce instance-level spatial quarantine, which reshapes the spatial belief landscape and suppresses repeated triggering.
Experiments on the HM3D benchmark show that \methodname{} consistently improves both Success Rate and Success weighted by Path Length, while substantially reducing false stops and cyclic deadlocks compared to prior frontier–VLM baselines. By transforming frontier exploration from scalar scoring into structured reasoning with probabilistic commitment, \methodname{} establishes a closed-loop decision paradigm for robust zero-shot object navigation. Code and models will be released.
\end{abstract}

\section{Introduction}
\begin{figure}[t]
\centering
\includegraphics[width=\columnwidth]{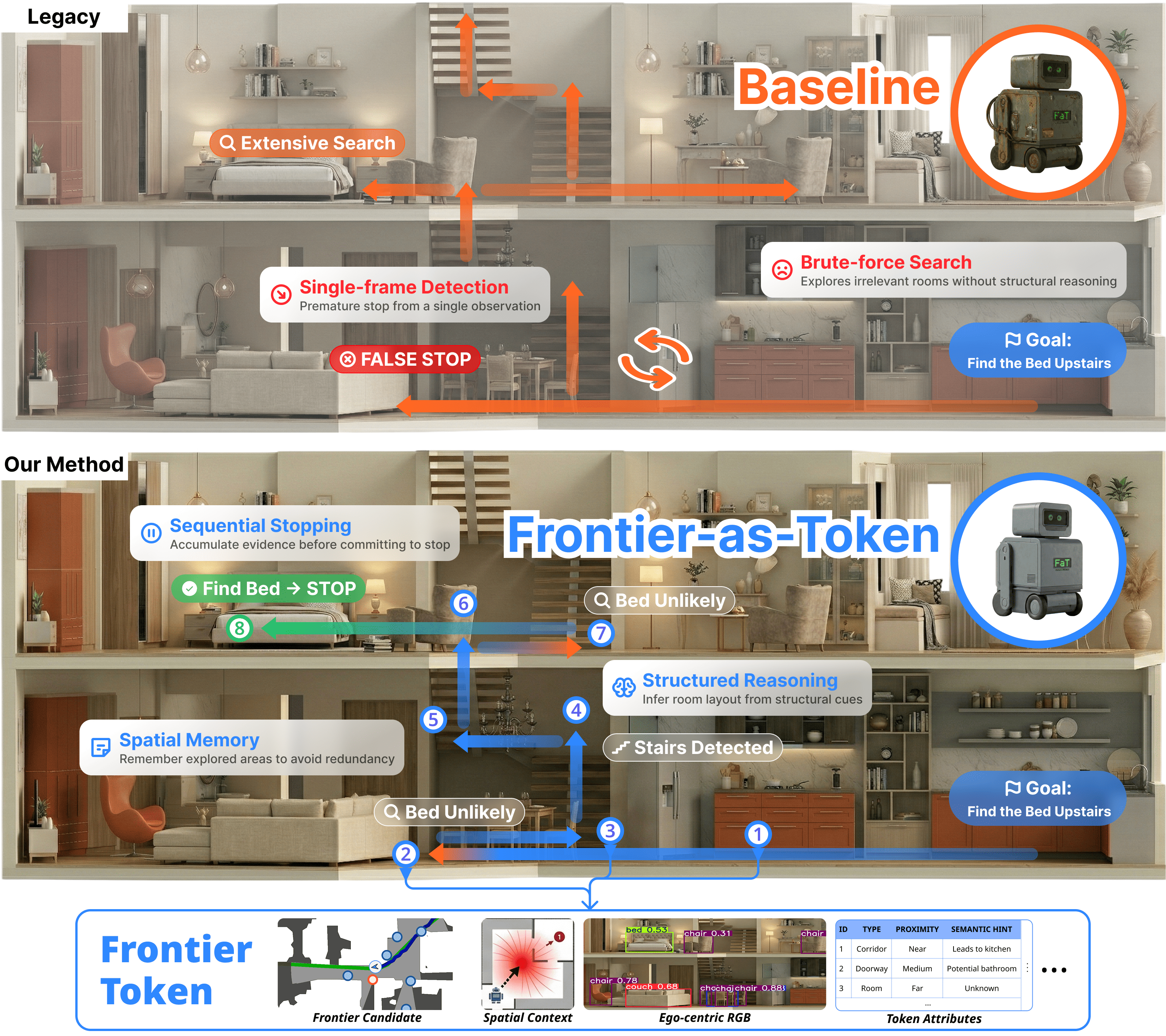}
\caption{\textbf{Frontier-as-Token for structured zero-shot ObjectNav.} Conventional frontier-based systems rely on brute-force exploration and single-frame detection, which often leads to inefficient search and premature stopping under ambiguous observations (top). In contrast, \methodname{} represents frontiers as persistent tokens and performs structured multimodal reasoning over spatial candidates (bottom). By combining spatial memory, comparative frontier reasoning, and sequential probabilistic stopping, the agent can infer scene structure and terminate only after sufficient evidence has been accumulated.}
\label{fig:teaser}
\vspace{-0.5\baselineskip}
\end{figure}

Object goal navigation (ObjectNav)~\cite{batra2020objectnav} requires an embodied agent to explore previously unseen indoor environments and locate a target object instance (e.g., \emph{toilet}) using only egocentric RGB-D observations. Beyond semantic recognition, ObjectNav fundamentally involves long-horizon spatial exploration and reliable termination: the agent must decide \emph{where to explore next} and \emph{when to stop}. These decisions become particularly challenging in zero-shot settings, where agents rely on foundation models to generalize across environments without task-specific training. Fig.~\ref{fig:teaser} illustrates both the failure modes of conventional frontier policies and the key components of \methodname{}.

Recent approaches incorporate vision-language models (VLMs) into frontier-based exploration~\cite{vlfm2023, gadre2023cow}, combining classical occupancy mapping with semantic scoring. While such methods benefit from strong open-vocabulary perception, they remain constrained by the underlying decision representation. In most frontier-based pipelines, spatial frontier regions are typically reduced to scalar scores and selected via an argmax rule. This reduction obscures geometric relationships between candidates, such as connectivity, room layout, and exploration potential, that are essential for comparative reasoning. Conversely, approaches that summarize spatial maps into textual abstractions~\cite{yokoyama2023esc, cognav2025} sacrifice geometric precision for linguistic reasoning, leading to a complementary loss of structural information. In both cases, the decision interface presented to the foundation model fails to preserve the spatial structure required for navigation.

Termination presents a second structural difficulty. Open-vocabulary perception models~\cite{clip-Radford2021, ditic_Zhou2022} can produce false positives, often in the form of persistent visual ambiguities (e.g., reflective surfaces resembling televisions). When stopping decisions rely on single-frame thresholds or heuristic recovery loops, agents either terminate prematurely or repeatedly revisit the same spurious instance, resulting in cyclic failures and deadlocks. Robust ObjectNav therefore requires not only improved perception, but also principled mechanisms for uncertainty-aware commitment and suppression of rejected hypotheses.

We address these limitations through \textbf{Frontier-as-Token (FaT)}, a unified decision framework that reformulates frontier exploration and stopping as structured reasoning and probabilistic commitment. Rather than treating frontiers as transient geometric points, FaT assigns persistent identities to frontier tokens that evolve consistently over time. At each decision step, the agent constructs a structured multimodal \emph{Set-of-Marks} representation consisting of (i) the current egocentric RGB observation with detections, (ii) a bird’s-eye occupancy map with indexed frontier tokens, and (iii) a feature table encoding geometric and semantic attributes. This representation enables a VLM to perform explicit comparative reasoning over spatially grounded candidates instead of operating on scalar summaries.

For termination, FaT replaces single-frame thresholding with a Bayesian Sequential Probability Ratio Test (SPRT), which accumulates evidence across time by integrating detection confidence, geometric consistency, and semantic verification. This probabilistic formulation explicitly decouples exploration from commitment and allows stopping decisions to emerge from sequential evidence rather than instantaneous scores. When a hypothesized target is rejected, FaT further introduces a 3D instance-level quarantine mechanism that projects a negative energy field into spatial memory around the rejected anchor~\cite{Khatib1985}. This reshaping of the belief landscape prevents repeated triggering of the same false instance and eliminates cyclic deadlocks without reliance on hand-crafted recovery behaviors.

We evaluate FaT on HM3D (v1 and v2) and MP3D under zero-shot settings. FaT achieves state-of-the-art performance across all benchmarks, reaching 79.1\% SR and 0.401 SPL on HM3Dv2 and 63.2\% SR / 0.348 SPL on HM3Dv1, while attaining the highest SPL (0.196) on MP3D. These improvements are achieved using a fully open-source VLM~\cite{qwen2025qwen3vlr}, while significantly reducing false stops and cyclic deadlocks. This shows that principled decision representation and uncertainty-aware commitment are critical for robust zero-shot object navigation.

In summary, \methodname{} makes three key contributions. (i) We reformulate frontier selection as token-level multimodal reasoning, preserving spatial structure for foundation models and enabling explicit comparative decisions. (ii) We introduce a sequential, uncertainty-aware stopping mechanism based on Bayesian evidence accumulation, providing principled commitment under open-vocabulary noise. (iii) We propose an energy-based 3D instance-level quarantine strategy that reshapes spatial belief to eliminate cyclic false detections without heuristic recovery routines.

\section{Related Work}
\label{sec:related}

Table~\ref{tab:comparison} summarizes the capability gap and highlights that FaT is the only framework that jointly supports structured candidate representation, explicit frontier reasoning, memory integration, and probabilistic commitment. 
 We will discuss related works along these dimensions below.

\noindent \textbf{Vision-Language Zero-Shot Object Navigation.}
Object goal navigation (ObjectNav)~\cite{batra2020objectnav} has evolved from environment-specific policy learning toward zero-shot paradigms that leverage large-scale vision-language models (VLMs) for semantic generalization. Earlier work explored semantic map-based navigation, where agents build spatial-semantic maps to guide object search (e.g., Semantic Exploration~\cite{Chaplot2020}). 
More recent approaches augment classical frontier exploration with vision-language models for semantic guidance and object localization~\cite{vlfm2023, gadre2023cow}. These methods introduce open-vocabulary perception but largely rely on scalar scores or geometric heuristics for frontier selection, limiting the model’s ability to reason over spatial relationships among alternatives. Complementarily, language-driven systems such as ESC~\cite{yokoyama2023esc}, SG-Nav~\cite{sgnav2024}, and CogNav~\cite{cognav2025} employ LLM-based reasoning over semantic and spatial abstractions. While enabling high-level inference, these approaches encode geometric structure indirectly through language, potentially sacrificing fine-grained layout information. In contrast, \methodname{} preserves geometric structure while enabling explicit multi-candidate reasoning through tokenized spatial representations.

\begin{table}[t]
\centering
\caption{Capability comparison of representative zero-shot ObjectNav methods.}
\label{tab:comparison}
\small
\setlength{\tabcolsep}{4pt}
\resizebox{1\linewidth}{!}{%
\begin{tabular}{lcccc}
\toprule
Method & Repr. & Compare & Memory & Commitment \\
\midrule
VLFM~\cite{vlfm2023}        & scalar &  &  &  \\
ESC~\cite{yokoyama2023esc}  & text   &  &  &  \\
ApexNav~\cite{apexnav2025}  & scalar &  & \checkmark &  \\
CogNav~\cite{cognav2025}    & text   &  &  &  \\
SG-Nav~\cite{sgnav2024}     & text   &  &  &  \\
RATE-Nav~\cite{ratenav2025} & text   &  &  &  \\
WMNav~\cite{nie2025wmnav}   & image  &  & \checkmark &  \\
\midrule
\textbf{FaT (Ours)} & \textbf{image+map+tokens} & \checkmark & \checkmark & \checkmark \\
\bottomrule
\end{tabular}}
\vspace{-2mm}
\end{table}

\begin{figure*}[t]
\centering
\includegraphics[width=\textwidth]{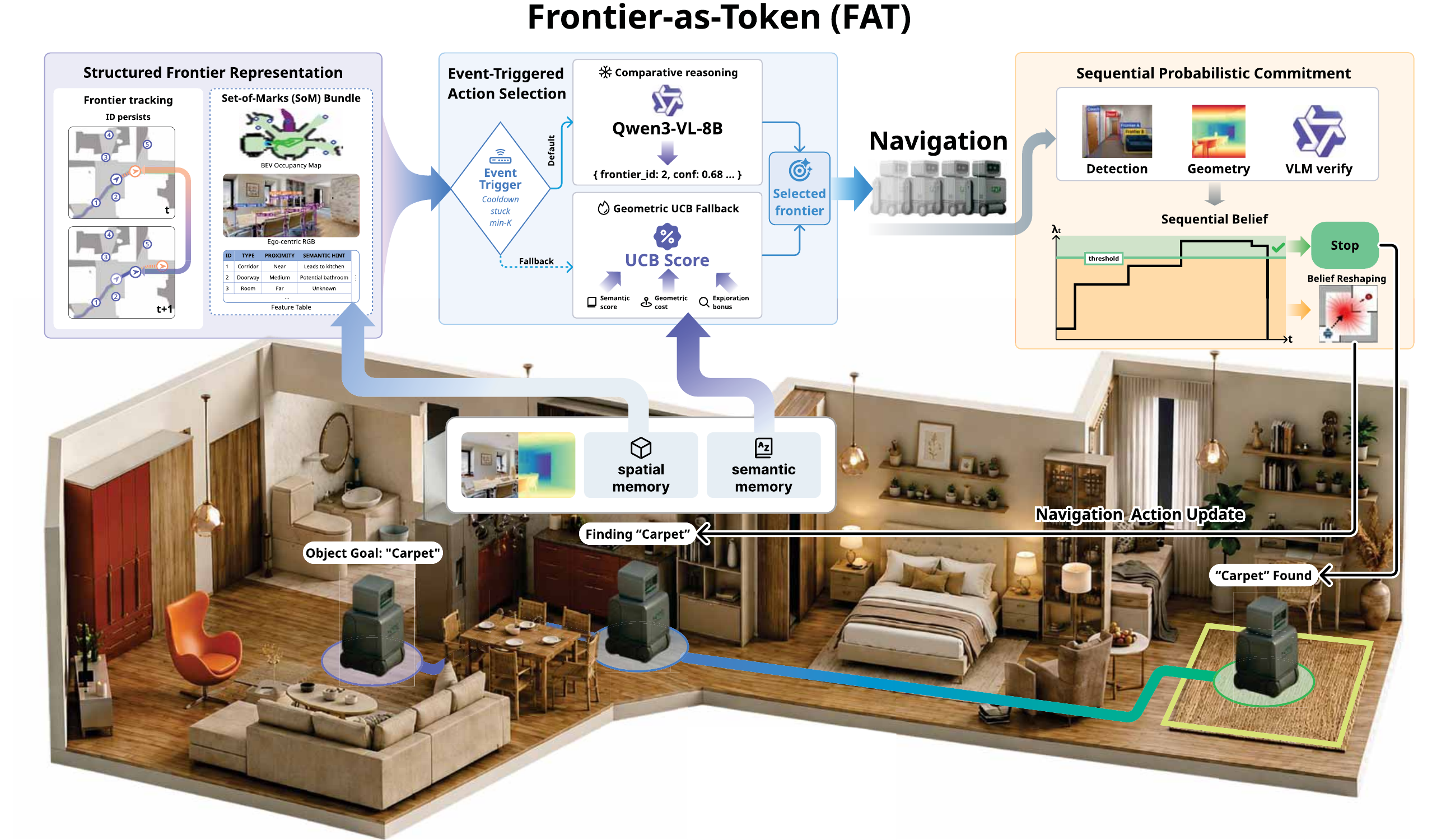}
\vspace{-6pt}
\caption{\textbf{FaT closed-loop architecture.} FaT represents each frontier as a persistent multimodal token and presents all candidates jointly via a Set-of-Marks (SoM) bundle (RGB with detections, bird’s-eye occupancy map with indexed frontiers, and a feature table). A VLM performs comparative reasoning for frontier selection, with a geometric UCB fallback to guarantee progress. Stopping is governed by Bayesian SPRT, and rejected hypotheses trigger a 3D instance-level quarantine that reshapes spatial belief to prevent cyclic failures.}
\vspace{-6pt}
\label{fig:architecture}
\end{figure*}

\vspace{4pt}
\noindent \textbf{Frontier-Based Exploration and Decision Representation.}
Frontier exploration~\cite{yamauchi1997frontier} remains a foundational strategy for mapping unknown environments. Classical methods prioritize frontiers using geometric heuristics such as distance~\cite{burgard2000collaborative}, information gain~\cite{stachniss2005information} or floor-plan prediction~\cite{song2025p}. Recent zero-shot variants incorporate semantic signals into exploration policies~\cite{vlfm2023, apexnav2025}. However, most frontier-based pipelines treat candidates independently and select actions via greedy maximization, implicitly assuming that spatial regions can be evaluated in isolation. Such scalar decision rules obscure relational structure between frontiers, including connectivity and contextual cues. \methodname{} departs from this formulation by representing frontiers as persistent tokens and performing structured multimodal reasoning over candidate sets rather than independent scoring.

\vspace{4pt}
\noindent \textbf{Commitment and Termination under Uncertainty.}
Reliable stopping remains a central challenge in ObjectNav, particularly under open-vocabulary detection noise. Many existing systems rely on single-frame confidence thresholds or temporal smoothing strategies~\cite{apexnav2025, ratenav2025}. Although multi-frame aggregation can improve stability, persistent ambiguities, such as reflections or visually similar objects, may still cause premature stopping or repeated revisitation.
To address this issue, we formulate stopping as sequential evidence accumulation using the Bayesian Sequential Probability Ratio Test (SPRT), rather than instantaneous thresholding. In addition, we introduce an instance-level spatial quarantine mechanism that reshapes the spatial belief associated with rejected hypotheses, preventing repeated false detections without relying on heuristic escape policies.

\vspace{4pt}
\noindent \textbf{Memory and Long-Horizon Navigation.}
Long-horizon embodied tasks require memory mechanisms to reduce redundant exploration. Prior work incorporates topological maps~\cite{toponav2025}, learned memory representations~\cite{efficientnav2025, ding2019camnet}, or language-conditioned recall~\cite{thinkremember2025, yuan2025remac}. These approaches primarily improve exploration efficiency by discouraging revisitation. FaT complements this direction by coupling memory with probabilistic commitment: instead of merely penalizing revisits, the proposed quarantine mechanism reshapes the spatial belief landscape, directly influencing future termination decisions.

\vspace{4pt}
\noindent\textbf{Positioning of FaT.}
Existing zero-shot ObjectNav systems typically (i) compress spatial candidates into scalar scores, (ii) abstract geometry into text-only reasoning, or (iii) rely on heuristic termination and recovery strategies. FaT unifies structured frontier reasoning, sequential probabilistic commitment, and belief shaping within a single closed-loop framework, providing a decision-centric perspective on robust zero-shot object navigation.

\section{Method}
\label{sec:method}

Frontier-as-Token (\methodname{}) follows a simple principle: robust zero-shot ObjectNav~\cite{batra2020objectnav} requires (i) a decision representation that preserves spatial structure and enables \emph{comparative} reasoning over frontier candidates, and (ii) stopping as \emph{probabilistic commitment} under open-vocabulary noise rather than single-frame thresholding. Following this principle, \methodname{} forms a closed-loop decision process that couples \emph{exploration} (where to go) with \emph{commitment} (when to stop). We present the framework following this loop: structured frontier representation and reasoning (Sec.~\ref{subsec:token}–\ref{subsec:som}), uncertainty-aware stopping via SPRT (Sec.~\ref{subsec:stop}), and belief shaping through instance-level quarantine (Sec.~\ref{subsec:quarantine}). The overall decision loop is summarized in Alg.~\ref{alg:fat_loop}.

\subsection{Problem Setup and Perception}
\label{subsec:setup}

At time $t$, the agent receives an egocentric RGB-D observation $o_t=(I_t,D_t)$. From these observations, the agent maintains an occupancy map $M_t$ and extracts geometric frontiers $\mathcal{F}_t=\{f_i\}$ defined as free–unknown boundaries~\cite{Thrun2005}. The agent executes an action $a_t\in\mathcal{A}$, consisting of motion primitives and a \texttt{STOP} action. The goal is to maximize navigation success and path efficiency while avoiding premature stopping and cyclic deadlocks. To support decision making under uncertainty, \methodname{} maintains three internal states: (i) a sequential stopping belief $\Lambda_t$, (ii) a spatial memory $\mathcal{S}_t$, and (iii) a semantic memory $\mathcal{E}_t$.

During exploration, we update the occupancy map from RGB-D observations and extract frontier candidates as connected free--unknown components. In addition, an open-vocabulary detector (YOLOE~\cite{wang2024yoloe}) provides target and co-occurrence detections with confidence scores, which are treated as noisy semantic cues for downstream reasoning. Together, these signals support the two key decisions in ObjectNav~\cite{batra2020objectnav}: selecting which frontier to explore next and determining when sufficient evidence has been accumulated to issue \texttt{STOP}. 
Fig.~\ref{fig:architecture} illustrates the overall decision pipeline.

\subsection{Tokenized Frontier Representation}
\label{subsec:token}

As the occupancy map evolves, frontier regions may appear, merge, split, or disappear across time steps. Instead of treating frontiers as transient geometric points, our \methodname{} assigns each frontier a persistent token identity so that spatial candidates can be tracked consistently across decisions. 

\begin{figure}[t]
\centering
\includegraphics[width=\columnwidth]{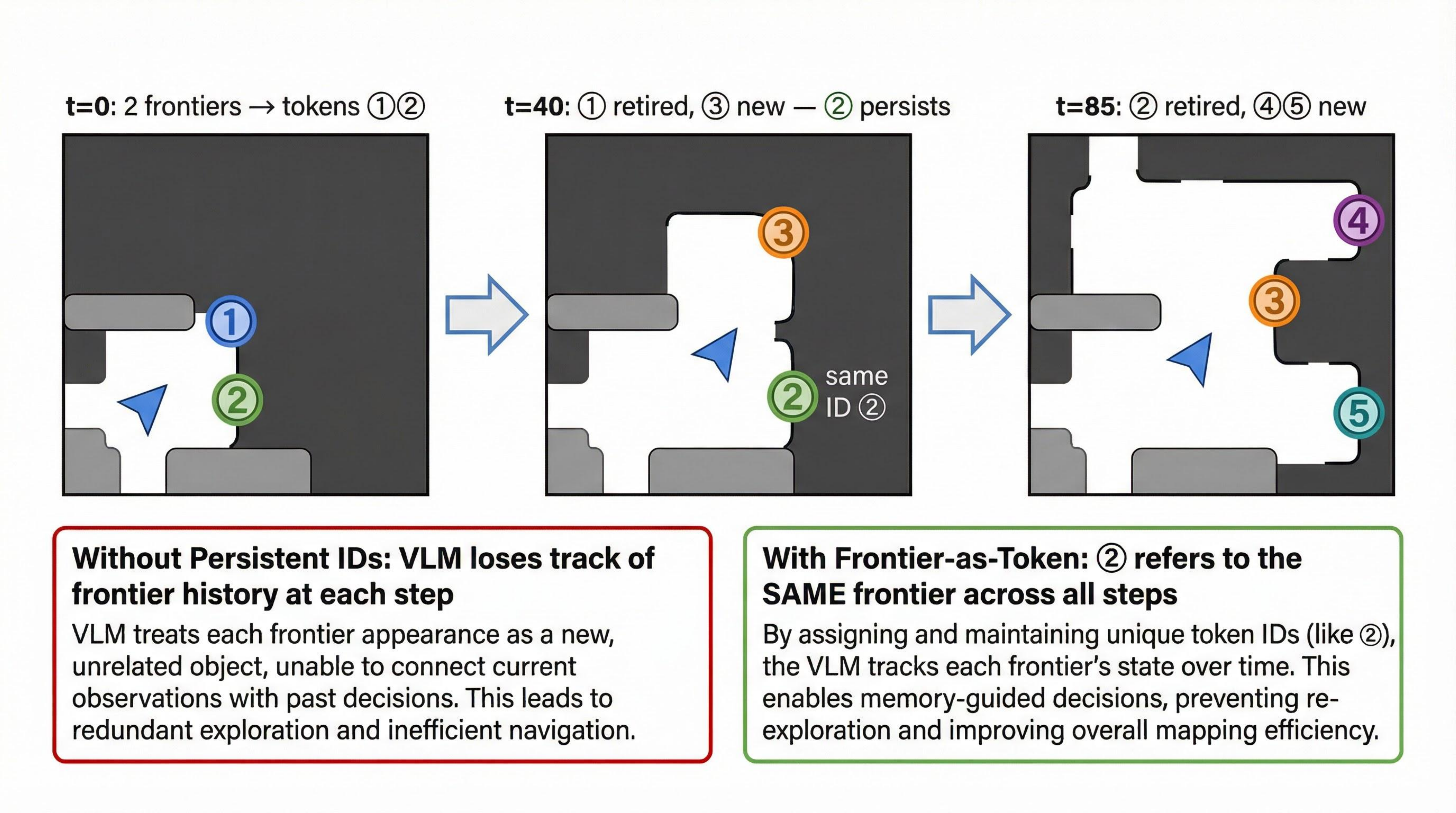}
\caption{\textbf{Persistent frontier tokens.} A frontier retains its ID across map evolution, enabling consistent reasoning about the same spatial candidate across multiple decision points.}
\label{fig:frontier_token}
\end{figure}

We maintain a \emph{FrontierTracker} that associates newly extracted frontier components with existing tokens using spatial nearest-neighbor matching. If a component matches an existing token, the token identity is preserved; otherwise a new token ID is created. Frontiers that are no longer observed in the updated map are removed. The resulting set of tokens at time $t$ is denoted by $\mathcal{T}_t=\{\tau_i\}$, corresponding to the frontier set $\mathcal{F}_t$. Each token $\tau_i$ stores its identity together with a set of attributes including geometric distance, semantic priors, memory signals, and selection history. This persistent representation enables the agent to reason about frontier candidates as stable entities rather than independent observations at each step. In particular, token identities allow contextual information to accumulate across time and prevent repeatedly selecting previously unproductive regions. The token set $\mathcal{T}_t$ forms the structured candidate interface used by the multimodal reasoning module described in Sec.~\ref{subsec:som}. Fig.~\ref{fig:frontier_token} illustrates how frontier tokens remain consistent across map updates.

\begin{figure}[t]
\centering
\includegraphics[width=\columnwidth]{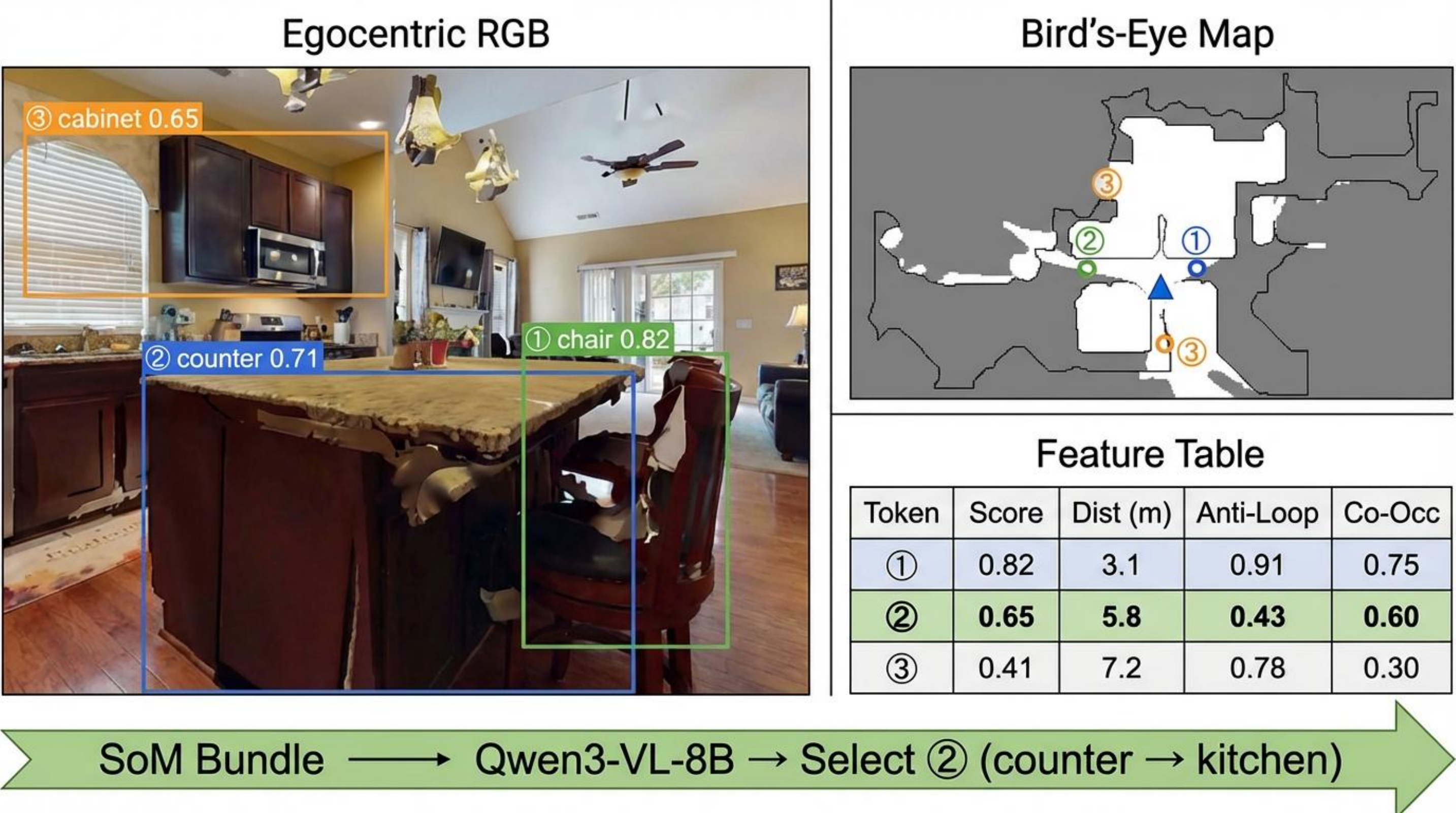}
\caption{\textbf{Set-of-Marks (SoM) bundle.} The VLM receives joint spatial context over all frontier tokens, enabling structured comparison grounded in geometry and semantic cues.}
\label{fig:som_bundle}
\end{figure}

\subsection{Structured Multimodal Candidate Reasoning}
\label{subsec:som}

To enable comparative reasoning over frontier tokens, \methodname{} presents the token set $\mathcal{T}_t$ to a vision–language model using a structured Set-of-Marks (SoM)~\cite{SOM2023} representation. The SoM bundle provides three complementary views of the navigation state: 
(i) the egocentric RGB observation annotated with object detections, (ii) a bird’s-eye occupancy map where frontier tokens are visualized with persistent indices, and (iii) a feature table describing each token with attributes such as geometric distance, semantic priors, and memory signals. By jointly presenting these elements, the SoM bundle preserves spatial structure while exposing semantic cues for navigation decisions. Fig.~\ref{fig:som_bundle} illustrates the resulting multimodal representation.

Given this structured input, the VLM performs comparative reasoning over frontier tokens and selects the most promising candidate for exploration. To balance reasoning quality and runtime efficiency, we employ the VLM in two complementary roles: a full multimodal reasoning module that operates on the complete SoM bundle during frontier selection, and a lightweight verification gate applied to cropped target observations during commitment. This decoupled design enables rich spatial reasoning without blocking real-time navigation.

When VLM confidence is low or reasoning is unavailable, we design a fallback to a geometric exploration policy based on Upper Confidence Bound (UCB)~\cite{auer2002finitetime}:

\begin{equation}
f^* = \arg\max_{f_i \in \mathcal{F}_t}
\left[\bar{s}_i + \beta \sqrt{\frac{\ln t}{n_i+1}}\right],
\end{equation}
where $\bar{s}_i$ denotes the running mean semantic score for frontier $f_i$, $n_i$ is the number of times $f_i$ has been selected, $t$ is the current decision step, and $\beta>0$ controls the exploration bonus. When multimodal reasoning is uncertain, this fallback guarantees continual progress and encourages occasional exploration of under-visited regions.

\subsection{Memory for Long-Horizon Exploration}
\label{subsec:memory}

To support long-horizon exploration under partial observability, our \methodname{} maintains lightweight spatial and semantic memories that provide historical context for frontier reasoning. These memories help prevent redundant exploration and improve candidate evaluation during navigation.

We create a \emph{SpatialMemoryBank} that indexes visited poses and produces repulsive and attractive signals that influence frontier scoring. Repulsive signals discourage revisiting explored regions, while attractive signals bias exploration toward unvisited areas. Short-term trajectory cycles are detected using sliding-window matching; when a cycle is detected, localized repulsive signals are injected to break deadlocks and encourage exploration of alternative frontiers.

In addition, a \emph{SemanticMemoryBank} stores observation embeddings in a FAISS index~\cite{Douze2024} for efficient vector similarity retrieval. Retrieved memories are injected into VLM prompts during frontier reasoning, allowing the agent to recall previously observed scene context. This enables more consistent semantic reasoning across exploration steps.

Together, spatial and semantic memories provide complementary signals that stabilize frontier selection and improve exploration efficiency in complex environments.

\subsection{Uncertainty-Aware Target Commitment}
\label{subsec:stop}

Termination is formulated as sequential probabilistic commitment rather than single-frame thresholding. \methodname{} maintains a scalar belief variable $\Lambda_t$, representing the accumulated log-odds of target presence. Following the principle of a Bayesian Sequential Probability Ratio Test (SPRT)~\cite{wald1947sequential}, the belief is updated as
\begin{align}
\Lambda_t = & \operatorname{max}\!\left(0,\Lambda_{t-1} +
\lambda_{\text{det}}(z_t) +
\lambda_{\text{geo}}(d_t) \right. \nonumber \\
& \left. + \lambda_{\text{vlm}}(v_t) +
\lambda_{\text{qua}}(p_t)\right),
\label{eq:lambda_update}
\end{align}
where $z_t$ denotes visual detection evidence (e.g., confidence and bounding box), $d_t$ represents depth-consistent proximity information, $v_t$ is a semantic verification signal provided by the VLM, and $p_t$ is the current hypothesized target location projected onto the occupancy map. The functions $\lambda_{\text{det}}(\cdot)$, $\lambda_{\text{geo}}(\cdot)$, and $\lambda_{\text{vlm}}(\cdot)$ convert these observations into incremental log-likelihood contributions that accumulate evidence over time, while $\lambda_{\text{qua}}(\cdot)\le 0$ provides a spatially localized negative correction induced by previously rejected hypotheses. The agent commits to \texttt{STOP} when $\Lambda_t \ge \Lambda_{\text{commit}}$, where $\Lambda_{\text{commit}}$ is a positive decision threshold.

To further reduce context-inconsistent detections, the commitment threshold is adaptively increased when expected co-occurrence cues are absent. This conservative adjustment improves robustness under ambiguous visual observations without disrupting the sequential evidence accumulation process.


\begin{algorithm}[t]
\SetAlgoLined
\KwIn{Step index $t$, observation $o_t$, target $g$, map $M_t$, frontiers $\mathcal{F}_t$, memories $\mathcal{S}_t,\mathcal{E}_t$, quarantine $\mathcal{Q}_t$}
\KwOut{Selected frontier $f^*$ or \texttt{STOP}}
\tcp{Commitment: sequential evidence accumulation (updates $\mathcal{Q}_t$ on rejection)}
\If{\textsc{UpdateSPRT}$(o_t,g,\mathcal{Q}_t)\ge \Lambda_{\text{commit}}$}{
  \Return \texttt{STOP}\;
}
\tcp{Exploration: tokenized candidates and structured reasoning}
$\mathcal{T}_t \leftarrow \textsc{TrackAndTokenize}(\mathcal{F}_t)$\;
\If{\textsc{TriggerCondition}()}{
  $B_t \leftarrow \textsc{BuildSoM}(o_t,M_t,\mathcal{T}_t,\mathcal{S}_t,\mathcal{E}_t)$\;
  $(f^*,c)\leftarrow \textsc{VLMSelect}(B_t,g)$\;
  \If{$c\ge\tau_c$}{
    \Return $f^*$\;
  }
}
\tcp{Robust fallback (UCB; statistics $\bar{s}_i$ and $n_i$ maintained per token)}
$f^* \leftarrow \arg\max_{f_i\in\mathcal{F}_t}\left[\bar{s}_i+\beta\sqrt{\ln t/(n_i+1)}\right]$\;
\Return $f^*$\;
\caption{FaT decision loop: structured exploration with probabilistic commitment}
\label{alg:fat_loop}
\end{algorithm}

\subsection{Belief Shaping via Instance-Level Quarantine}
\label{subsec:quarantine}

Even with sequential evidence accumulation, persistent visual ambiguities (such as reflective surfaces or visually similar objects) may still trigger repeated false commitments. To prevent such cyclic failures, we reshape the spatial belief landscape whenever a target hypothesis is rejected.

Specifically, when a hypothesized target anchored at 3D location $A_k$ is rejected, we project a negative Gaussian energy field centered at its bird's-eye projection $A_k^{\text{bev}}$:
\begin{equation}
\mathcal{E}_k(p) =
-\gamma \exp\!\left(-\frac{\|p - A_k^{\text{bev}}\|^2}{2\sigma_q^2}\right),
\end{equation}
where $p$ denotes a 2D position on the occupancy map, $\gamma>0$ controls the suppression strength, and $\sigma_q$ determines the spatial extent of the quarantined region.

The quarantine correction used in Eq.~\eqref{eq:lambda_update} is obtained by evaluating the accumulated quarantine energy at the current hypothesis location:
\begin{equation}
\lambda_{\text{qua}}(p_t) = \sum_{k \in \mathcal{Q}_t} \mathcal{E}_k(p_t),
\end{equation}
where $p_t$ denotes the current hypothesis location on the occupancy map, and $\mathcal{Q}_t$ denotes the set of rejected target anchors up to time $t$. As a result, if the agent revisits a region near a previously rejected hypothesis, the local quarantine field contributes a negative term to the belief update and suppresses repeated commitment to the same false instance.

Rather than relying on heuristic recovery actions, our \methodname{} modifies the spatial belief landscape directly. This prevents repeated commitment to the same false instance and naturally redirects exploration toward alternative frontier tokens.
\subsection{Overall Decision Loop}
\label{subsec:overall}

Alg.~\ref{alg:fat_loop} summarizes our \methodname{} decision architecture. At each step, the agent first evaluates the stopping belief through sequential evidence accumulation. Exploration proceeds only if the belief remains below the commitment threshold. Frontier selection is performed over tokenized candidates using event-triggered VLM reasoning. When reasoning confidence is low or unavailable, a geometric UCB policy serves as a fallback to ensure continual progress under uncertainty. If a target hypothesis is rejected during verification, the associated anchor is quarantined, reshaping the spatial belief landscape and preventing cyclic re-commitment. Together, structured exploration, probabilistic commitment, and belief shaping form the closed-loop decision process underlying \methodname{}.

\section{Experiments}
\label{sec:experiments}

We evaluate Frontier-as-Token (FaT) on three standard zero-shot ObjectNav benchmarks. Our experiments are organized to validate three claims aligned with the method design: (i) structured frontier reasoning improves long-horizon exploration compared to scalar frontier scoring, (ii) sequential commitment reduces false stops and cyclic deadlocks under open-vocabulary noise, and (iii) FaT achieves competitive or state-of-the-art performance against prior systems.

\begin{figure}[t]
\centering
\includegraphics[width=\columnwidth]{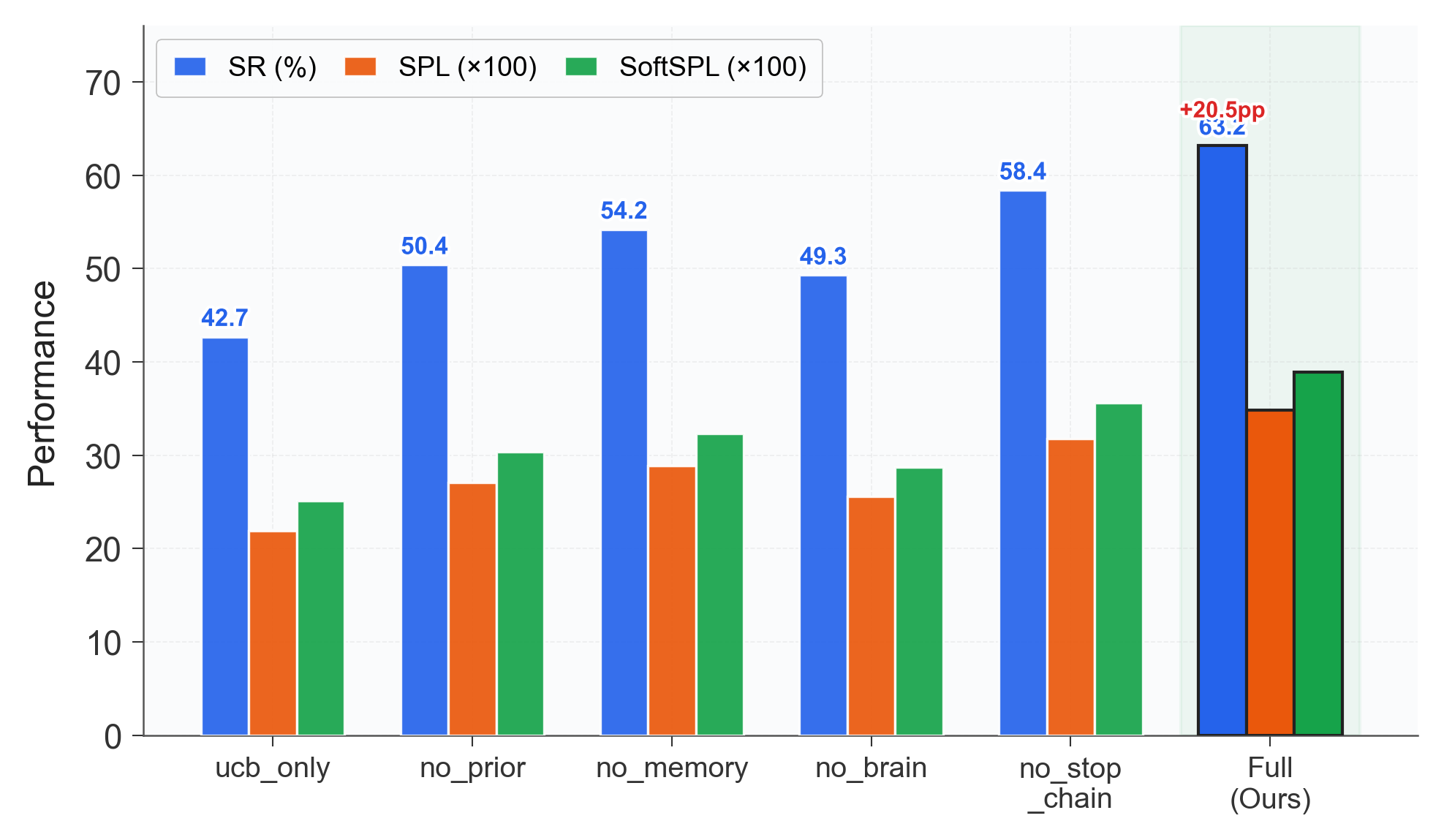}
\caption{Ablation on HM3Dv1 validation. The full system achieves 63.2\% SR. Removing VLM frontier reasoning (\texttt{no\_brain}) yields a 13.9\,pp SR drop; removing priors (\texttt{no\_prior}) costs 12.8\,pp; removing memory (\texttt{no\_memory}) costs 9.0\,pp; replacing sequential commitment with single-frame stopping (\texttt{no\_stop\_chain}) costs 4.8\,pp.}
\label{fig:ablation_chart}
\end{figure}

\subsection{Experimental Setup}

\vspace{4pt}
\noindent\textbf{Datasets.}
We evaluate \methodname{} on three standard ObjectNav benchmarks:
HM3Dv1~\cite{ramakrishnan2021hm3d} (6 categories, 20 validation scenes, 2000 episodes),
HM3Dv2~\cite{yadav2023ovrl} (6 categories, 36 validation scenes, 1000 episodes),
and MP3D~\cite{chang2017mp3d} (21 categories, 11 validation scenes, 2195 episodes).

\vspace{4pt}
\noindent\textbf{Metrics.} We report Success Rate (SR) and Success weighted by Path Length (SPL)~\cite{anderson2018spl}. Following standard protocols~\cite{habitatchallenge2022}, an episode succeeds if the agent calls \texttt{STOP} within 1.0\,m of the target for HM3D (0.2\,m for MP3D) within a budget of $T_{\max}=500$ steps.

\vspace{4pt}
\noindent\textbf{Implementation details.} We use YOLOE~\cite{wang2024yoloe} (0.4 confidence threshold) for open-vocabulary detection and Qwen3-VL-8B~\cite{qwen2025qwen3vlr} for VLM reasoning. The agent builds a 5\,cm/cell occupancy map and uses spatial memory parameters $\sigma_r{=}2.0$\,m and $\lambda_a{=}3.0$\,m, with a UCB fallback parameter $\beta{=}1.4$. VLM inference runs asynchronously such that exploration proceeds during inference. All results are deterministic (seed=0, temperature=0). Complete implementation details and hyperparameters are provided in the released codebase.

\subsection{Ablation Study (HM3Dv1)}

We ablate FaT to quantify the contribution of each component to exploration and commitment (Table~\ref{tab:main_results}, Fig.~\ref{fig:ablation_chart}). Variants range from a geometric-only baseline (\texttt{ucb\_only}) to the \textbf{Full} system.

\begin{table}[t]
\centering
\caption{Ablation results on the HM3Dv1 validation set. Best in \textbf{bold}. $\Delta$ indicates absolute changes relative to the \texttt{ucb\_only} geometric baseline.}
\label{tab:main_results}
\begin{tabular}{lcccc}
\toprule
\textbf{Method} & \textbf{SR$\uparrow$} & \textbf{SPL$\uparrow$} & \textbf{DtG$\downarrow$} & \textbf{SoftSPL$\uparrow$} \\
\midrule
ucb\_only         & 42.7 & 0.219 & 3.21 & 0.251 \\
no\_prior         & 50.4 & 0.261 & 2.85 & 0.298 \\
no\_memory        & 54.2 & 0.289 & 2.63 & 0.323 \\
no\_brain         & 49.3 & 0.256 & 2.91 & 0.287 \\
no\_stop\_chain   & 58.4 & 0.312 & 2.34 & 0.354 \\
\textbf{Full}     & \textbf{63.2} & \textbf{0.348} & \textbf{2.07} & \textbf{0.389} \\
\midrule
$\Delta$ (Full $-$ ucb\_only) & +20.5 & +0.129 & $-$1.14 & +0.138 \\
\bottomrule
\end{tabular}
\vspace{6pt}
\end{table}

\begin{table}[t]
\centering
\caption{Comparison with zero-shot ObjectNav methods. ``---'' = not reported. $\star$ = best SPL on HM3Dv1.}
\label{tab:sota_comparison}
\setlength{\tabcolsep}{6pt}
\small
\begin{tabular}{l cc cc cc}
\toprule
\multirow{2}{*}{\textbf{Method}}
  & \multicolumn{2}{c}{\textbf{HM3Dv1}} & \multicolumn{2}{c}{\textbf{HM3Dv2}} & \multicolumn{2}{c}{\textbf{MP3D}} \\
\cmidrule(lr){2-3}\cmidrule(lr){4-5}\cmidrule(lr){6-7}
 & SR & SPL & SR & SPL & SR & SPL \\
\midrule
ESC~\cite{yokoyama2023esc}       & 39.2 & .223 & ---  & ---  & 28.7 & .142 \\
L3MVN~\cite{l3mvn2023}           & 50.4 & .231 & 36.3 & .157 & 34.9 & .145 \\
VLFM~\cite{vlfm2023}             & 52.5 & .304 & 63.6 & .325 & 36.4 & .175 \\
SG-Nav~\cite{sgnav2024}          & 54.0 & .249 & 49.6 & .255 & 40.2 & .160 \\
ApexNav~\cite{apexnav2025}       & 59.6 & .330 & 76.2 & .380 & 39.2 & .178 \\
WMNav~\cite{nie2025wmnav}        & 58.1 & .312 & ---  & ---  & 45.4 & .172 \\
\midrule
\textbf{Ours (FaT)}
  & \textbf{63.2} & $\star$\textbf{.348}
  & \textbf{79.1} & \textbf{.401}
  & 41.8 & \textbf{.196} \\
\bottomrule
\end{tabular}
\end{table}

\vspace{4pt}
\noindent\textbf{Structured frontier reasoning drives efficiency.}
Disabling VLM token reasoning (\texttt{no\_brain}) reduces SR by 13.9\,pp and SPL by 0.092, indicating that scalar frontier scoring is insufficient to capture spatial-semantic structure for long-horizon exploration.

\vspace{4pt}
\noindent\textbf{Priors and memory accelerate discovery.}
Removing semantic priors (\texttt{no\_prior}) incurs a 12.8\,pp SR penalty, while disabling spatial and semantic memory (\texttt{no\_memory}) reduces SR by 9.0\,pp. These results suggest that contextual cues and long-horizon recall significantly reduce redundant exploration.

\vspace{4pt}
\noindent\textbf{Sequential commitment improves robustness.}
Replacing SPRT-based commitment and quarantine with single-frame stopping (\texttt{no\_stop\_chain}) reduces SR by 4.8\,pp, consistent with fewer premature stops and fewer cyclic deadlocks under uncertainty.

\subsection{Comparison with State-of-the-Art}

Table~\ref{tab:sota_comparison} compares FaT against leading zero-shot ObjectNav methods. FaT achieves the highest SR on both HM3D benchmarks and the best SPL across all datasets, while using a fully open-source VLM (Qwen3-VL-8B)~\cite{qwen2025qwen3vlr}. On HM3Dv2, FaT reaches 79.1\% SR and 0.401 SPL, improving over ApexNav~\cite{apexnav2025} by +2.9\,pp SR. On MP3D, FaT attains the highest SPL (0.196), indicating superior path efficiency.

\subsection{Qualitative and Failure Analysis}

\begin{figure}[t]
\centering
\includegraphics[width=\columnwidth]{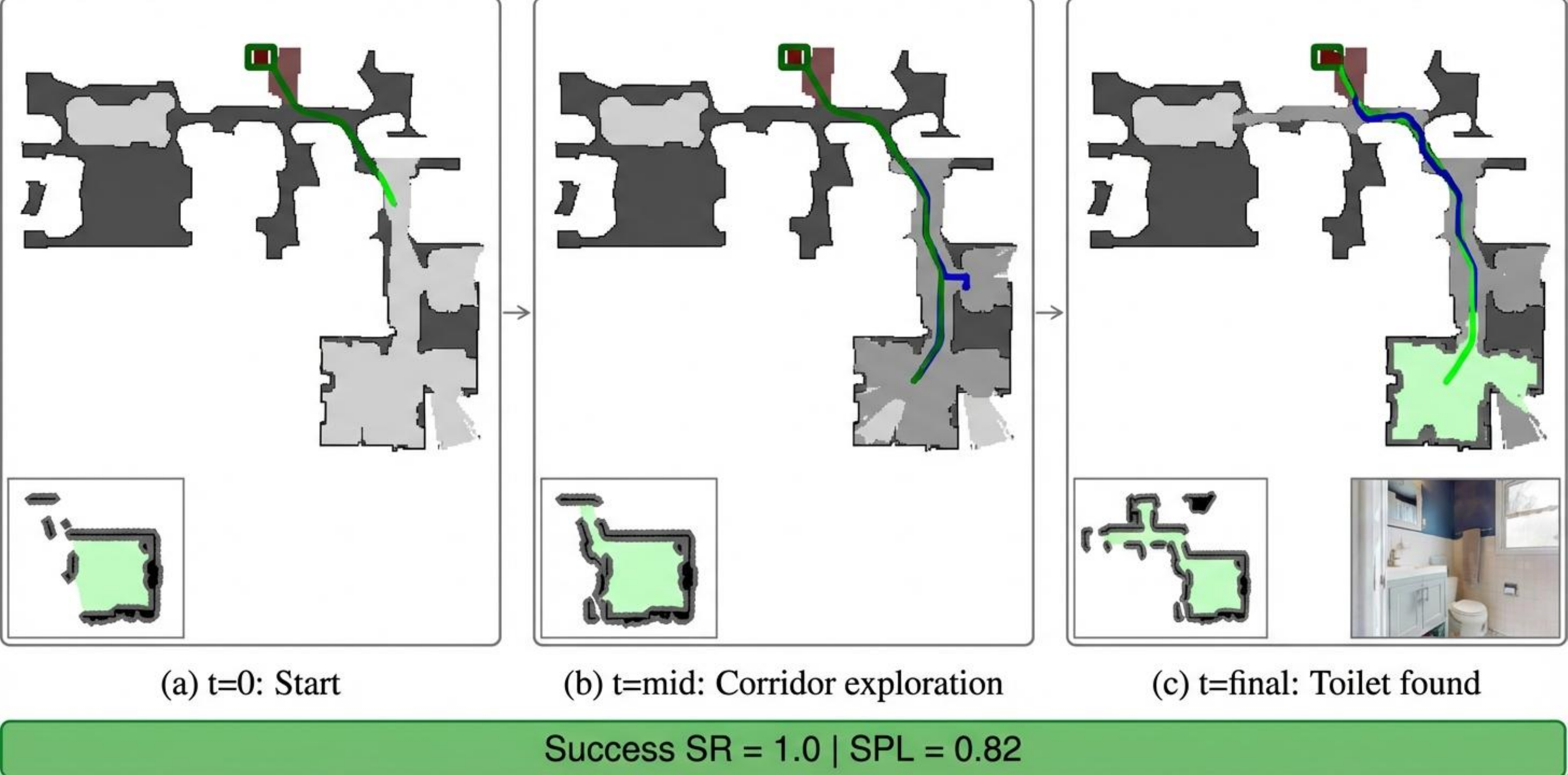}
\caption{Qualitative HM3D episode (target: toilet). Bird’s-eye maps show the trajectory and frontier candidates over time; the agent progressively expands coverage and terminates successfully (SPL=0.82).}
\label{fig:qualitative}
\end{figure}


Fig.~\ref{fig:qualitative} illustrates a representative successful episode. Beyond successes, we observe three recurring failure modes.

\begin{figure}[t]
\centering
\includegraphics[width=\linewidth]{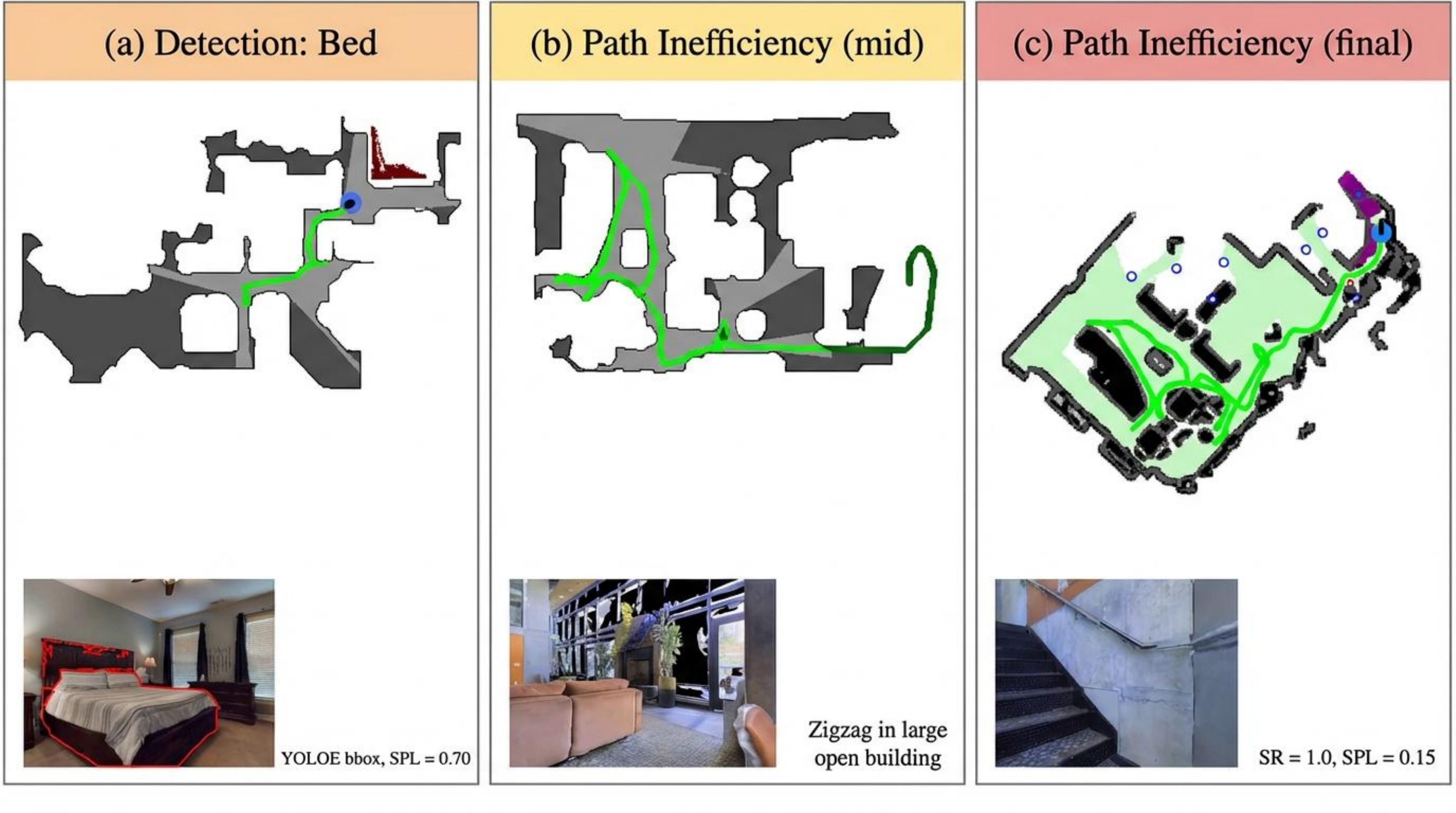}
\caption{Failure modes. (a) Example of reliable detection; (b,c) path inefficiency in open scenes due to repeated revisitation. Large open layouts remain challenging for frontier-based exploration.}
\label{fig:failure_cases}
\end{figure}

\vspace{4pt}
\noindent \textbf{(1) Path inefficiency in large open scenes} (Fig.~\ref{fig:failure_cases} (b,c)).
When the environment is open-plan, the frontier set grows rapidly and long-range prioritization becomes difficult, leading to zigzag revisitation patterns and reduced SPL even when SR remains high.

\vspace{4pt}
\noindent \textbf{(2) Visually ambiguous objects (wrong-object stops).}
Some categories exhibit strong visual aliasing (e.g., couches vs.\ beds), which can still trigger premature commitment in rare cases despite sequential evidence accumulation.

\vspace{4pt}
\noindent \textbf{(3) Near-target timeouts.}
For structurally complex or occluded targets, the agent may approach within proximity but fail to accumulate sufficient evidence to commit before exhausting the step budget.

These observations suggest that structured frontier reasoning and probabilistic commitment address complementary failure modes in zero-shot ObjectNav.

\subsection{Computational Overhead}

FaT invokes Qwen3-VL only on discrete trigger events, averaging $\approx$52 calls per episode. Compared with the geometric baseline (\texttt{ucb\_only}), this introduces additional runtime on an A100 (231\,s vs.\ 148\,s). However, the overhead remains practical under asynchronous deployment: while VLM reasoning is executed, the agent continues local exploration using the UCB fallback policy. YOLOE operates at real-time speed ($<$60\,ms/frame), preserving closed-loop responsiveness during final approach and obstacle avoidance.

\subsection{Limitations and Future Directions} 
\methodname{} relies on open-vocabulary detection to provide semantic cues for reasoning and commitment. When a target category is weakly represented in the detector’s embedding space, the available visual evidence may be limited. Rather than being specific to our framework, this limitation reflects a broader challenge in foundation-model-based navigation: robust semantic grounding remains a key bottleneck for zero-shot embodied agents.

In addition, environments with large open-plan layouts and weak structural cues can reduce path efficiency, as frontier-based exploration provides fewer geometric signals for prioritizing distant unexplored regions. Interestingly, such cases highlight opportunities for future research on hierarchical exploration strategies and long-range spatial reasoning. We believe that combining structured decision representations, as introduced in \methodname{}, with stronger grounding models and higher-level planning mechanisms represents a promising direction for building more capable and reliable embodied navigation systems.

\section{Conclusion}
\label{sec:conclusion}

We presented Frontier-as-Token (\methodname{}), a unified decision framework for zero-shot object navigation that explicitly aligns spatial representations with foundation-model reasoning. Instead of collapsing frontier candidates into scalar scores, \methodname{} represents geometric frontiers as persistent multimodal tokens and enables structured comparative reasoning over spatial alternatives, preserving spatial context and exposing rich relational cues for decision making. 
Building on this representation, \methodname{} introduces uncertainty-aware commitment through Bayesian SPRT and belief shaping via instance-level spatial quarantine, allowing agents to accumulate evidence over time and avoid cyclic failures under open-vocabulary uncertainty. Together, these components transform frontier exploration from heuristic scoring into a principled closed-loop decision process that integrates spatial memory, semantic reasoning, and probabilistic commitment.

Across HM3D and MP3D benchmarks, \methodname{} achieves state-of-the-art Success Rates and the best SPL among zero-shot navigation systems while using a fully open-source VLM. Beyond empirical performance, our results highlight a broader insight: improving the \emph{decision structure} of embodied agents through structured representations and uncertainty-aware reasoning can significantly enhance robustness, even without scaling perception models. We hope this perspective encourages future research on decision-centric architectures for embodied intelligence and provides a foundation for integrating foundation models with spatial reasoning in interactive environments.

{
\bibliographystyle{IEEEtran}
\bibliography{references}

@inproceedings{vlfm2023,
  title     = {{VLFM}: Vision-Language Frontier Maps for Zero-Shot Semantic Navigation},
  author    = {Yokoyama, Naoki and Ha, Sehoon and Batra, Dhruv and Wang, Jiuguang and Bucher, Bernadette},
  booktitle = {Proceedings of the IEEE International Conference on Robotics and Automation (ICRA)},
  year      = {2024}
}

@inproceedings{gadre2023cow,
  title     = {{CoWs on Pasture}: Baselines and Benchmarks for Language-Driven Zero-Shot Object Navigation},
  author    = {Gadre, Samir Yitzhak and Wortsman, Mitchell and Ilharco, Gabriel and Schmidt, Ludwig and Song, Shuran},
  booktitle = {Proceedings of the IEEE/CVF Conference on Computer Vision and Pattern Recognition (CVPR)},
  pages     = {23171--23181},
  year      = {2023}
}

@article{yuan2025remac,
  title={Remac: Self-reflective and self-evolving multi-agent collaboration for long-horizon robot manipulation},
  author={Yuan, Puzhen and Ma, Angyuan and Yao, Yunchao and Yao, Huaxiu and Tomizuka, Masayoshi and Ding, Mingyu},
  journal={arXiv preprint arXiv:2503.22122},
  year={2025}
}

@inproceedings{song2025p,
  title={P2 Explore: Efficient Exploration in Unknown Cluttered Environment with Floor Plan Prediction},
  author={Song, Kun and Chen, Gaoming and Tomizuka, Masayoshi and Zhan, Wei and Xiong, Zhenhua and Ding, Mingyu},
  booktitle={2025 IEEE/RSJ International Conference on Intelligent Robots and Systems (IROS)},
  pages={13090--13096},
  year={2025},
  organization={IEEE}
}

@inproceedings{ding2019camnet,
  title={CamNet: Coarse-to-fine retrieval for camera re-localization},
  author={Ding, Mingyu and Wang, Zhe and Sun, Jiankai and Shi, Jianping and Luo, Ping},
  booktitle={Proceedings of the IEEE/CVF International Conference on Computer Vision},
  pages={2871--2880},
  year={2019}
}

@inproceedings{yamauchi1997frontier,
  title     = {A Frontier-Based Approach for Autonomous Exploration},
  author    = {Yamauchi, Brian},
  booktitle = {Proceedings of the 1997 IEEE International Symposium on Computational Intelligence in Robotics and Automation (CIRA)},
  pages     = {146--151},
  year      = {1997},
  publisher = {IEEE},
  address   = {Monterey, CA, USA},
  doi       = {10.1109/CIRA.1997.613851}
}

@inproceedings{burgard2000collaborative,
  title     = {Collaborative Multi-Robot Exploration},
  author    = {Burgard, Wolfram and Moors, Mark and Fox, Dieter and Simmons, Reid and Thrun, Sebastian},
  booktitle = {Proceedings of the IEEE International Conference on Robotics and Automation (ICRA)},
  volume    = {1},
  pages     = {476--481},
  year      = {2000},
  organization = {IEEE}
}

@inproceedings{stachniss2005information,
  title     = {Information Gain-based Exploration Using {Rao-Blackwellized} Particle Filters},
  author    = {Stachniss, Cyrill and Grisetti, Giorgio and Burgard, Wolfram},
  booktitle = {Proceedings of Robotics: Science and Systems (RSS)},
  year      = {2005},
  address   = {Cambridge, MA, USA}
}

@article{batra2020objectnav,
  title     = {{ObjectNav Revisited}: On Evaluation of Embodied Agents Navigating to Objects},
  author    = {Batra, Dhruv and Gokaslan, Aaron and Kembhavi, Aniruddha and Maksymets, Oleksandr and Mottaghi, Roozbeh and Savva, Manolis and Toshev, Alexander and Wijmans, Erik},
  journal   = {arXiv preprint arXiv:2006.13171},
  year      = {2020}
}

@inproceedings{ramakrishnan2021hm3d,
  title     = {Habitat-Matterport {3D} Dataset ({HM3D}): 1000 Large-scale {3D} Environments for Embodied {AI}},
  author    = {Ramakrishnan, Santhosh Kumar and Gokaslan, Aaron and Wijmans, Erik and Maksymets, Oleksandr and Clegg, Alexander and Turner, John and Undersander, Eric and Galuba, Wojciech and Westbury, Andrew and Chang, Angel X. and Savva, Manolis and Zhao, Yili and Batra, Dhruv},
  booktitle = {Thirty-fifth Conference on Neural Information Processing Systems (NeurIPS) Datasets and Benchmarks Track},
  year      = {2021}
}

@misc{habitatchallenge2022,
  title        = {Habitat Challenge 2022},
  author       = {{Habitat Team}},
  howpublished = {\url{https://aihabitat.org/challenge/2022/}},
  year         = {2022}
}

@article{wang2024yoloe,
  title     = {{YOLOE}: Real-Time Seeing Anything},
  author    = {Wang, Ao and Liu, Lihao and Chen, Hui and Lin, Zijia and Han, Jungong and Ding, Guiguang},
  journal   = {arXiv preprint arXiv:2503.07465},
  year      = {2025},
}

@article{qwen2025qwen3vlr,
  title     = {Qwen3-VL Technical Report},
  author    = {{Qwen Team}},
  journal   = {arXiv preprint arXiv:2511.21631},
  year      = {2025},
}

@inproceedings{sgnav2024,
  title     = {{SG-Nav}: Online 3D Scene Graph Prompting for {LLM}-based Zero-Shot Object Navigation},
  author    = {Yin, Hang and Xu, Xiuwei and Wu, Zhenyu and Zhou, Jie and Lu, Jiwen},
  booktitle = {Advances in Neural Information Processing Systems (NeurIPS)},
  year      = {2024},
  url       = {https://arxiv.org/abs/2410.08189},
}

@article{apexnav2025,
  title     = {{ApexNav}: An Adaptive Exploration Strategy for Zero-Shot Object Navigation with Target-centric Semantic Fusion},
  author    = {Zhang, Mingjie and Du, Yuheng and Wu, Chengkai and Zhou, Jinni and Qi, Zhenchao and Ma, Jun and Zhou, Boyu},
  journal   = {IEEE Robotics and Automation Letters},
  year      = {2025},
}

@article{toponav2025,
  title     = {{TopoNav}: Topological Graphs as a Key Enabler for Advanced Object Navigation},
  author    = {Liu, Peiran and Zhang, Qiang and Peng, Daojie and Zhang, Lingfeng and Qin, Yihao and Zhou, Hang and Ma, Jun and Xu, Renjing and Ji, Yiding},
  journal   = {arXiv preprint arXiv:2509.01364},
  year      = {2025},
}

@article{efficientnav2025,
  title     = {{EfficientNav}: Towards On-Device Object-Goal Navigation with Navigation Map Caching and Retrieval},
  author    = {Yang, Zebin and Zheng, Sunjian and Xie, Tong and Xu, Tianshi and Yu, Bo and Wang, Fan and Tang, Jie and Liu, Shaoshan and Li, Meng},
  journal   = {arXiv preprint arXiv:2510.18546},
  year      = {2025},
}

@article{thinkremember2025,
  title     = {Think, Remember, Navigate: Zero-Shot Object-Goal Navigation with {VLM}-Powered Reasoning},
  author    = {Habibpour, Mobin and Afghah, Fatemeh},
  journal   = {arXiv preprint arXiv:2511.08942},
  year      = {2025},
}

@inproceedings{l3mvn2023,
  title     = {{L3MVN}: Leveraging Large Language Models for Visual Target Navigation},
  author    = {Yu, Bangguo and Kasaei, Hamidreza and Cao, Ming},
  booktitle = {2023 IEEE/RSJ International Conference on Intelligent Robots and Systems (IROS)},
  pages     = {3554--3560},
  year      = {2023},
  organization = {IEEE}
}

@article{chang2017mp3d,
  title     = {Matterport3D: Learning from {RGB-D} Data in Indoor Environments},
  author    = {Chang, Angel and Dai, Angela and Funkhouser, Thomas and Halber, Maciej and Niessner, Matthias and Savva, Manolis and Song, Shuran and Zeng, Andy and Zhang, Yinda},
  journal   = {arXiv preprint arXiv:1709.06158},
  year      = {2017}
}

@book{wald1947sequential,
  title={Sequential Analysis},
  author={Wald, Abraham},
  year={1947},
  publisher={John Wiley \& Sons},
  address={New York}
}

@article{auer2002finitetime,
  title={Finite-time Analysis of the Multiarmed Bandit Problem},
  author={Auer, Peter and Cesa-Bianchi, Nicolo and Fischer, Paul},
  journal={Machine Learning},
  volume={47},
  number={2-3},
  pages={235--256},
  year={2002},
  publisher={Springer}
}

@Misc{clip-Radford2021,
  author        = {Alec Radford and Jong Wook Kim and Chris Hallacy and Aditya Ramesh and Gabriel Goh and Sandhini Agarwal and Girish Sastry and Amanda Askell and Pamela Mishkin and Jack Clark and Gretchen Krueger and Ilya Sutskever},
  title         = {Learning Transferable Visual Models From Natural Language Supervision},
  year          = {2021},
  archiveprefix = {arXiv},
  eprint        = {2103.00020},
  primaryclass  = {cs.CV},
  url           = {https://arxiv.org/abs/2103.00020},
}

@Misc{ditic_Zhou2022,
  author        = {Xingyi Zhou and Rohit Girdhar and Armand Joulin and Philipp Krähenbühl and Ishan Misra},
  title         = {Detecting Twenty-thousand Classes using Image-level Supervision},
  year          = {2022},
  archiveprefix = {arXiv},
  eprint        = {2201.02605},
  primaryclass  = {cs.CV},
  url           = {https://arxiv.org/abs/2201.02605},
}

@Book{Thrun2005,
  author    = {Thrun, Sebastian and Burgard, Wolfram and Fox, Dieter},
  publisher = {The MIT Press},
  title     = {Probabilistic Robotics (Intelligent Robotics and Autonomous Agents)},
  year      = {2005},
  isbn      = {0262201623},
}

@InProceedings{Khatib1985,
  author    = {Khatib, O.},
  booktitle = {Proceedings. 1985 IEEE International Conference on Robotics and Automation},
  title     = {Real-time obstacle avoidance for manipulators and mobile robots},
  year      = {1985},
  pages     = {500-505},
  volume    = {2},
  doi       = {10.1109/ROBOT.1985.1087247},
}

@InProceedings{Chaplot2020,
  author    = {Chaplot, Devendra Singh and Gandhi, Dhiraj Prakashchand and Gupta, Abhinav and Salakhutdinov, Russ R},
  booktitle = {Advances in Neural Information Processing Systems},
  title     = {Object Goal Navigation using Goal-Oriented Semantic Exploration},
  year      = {2020},
  editor    = {H. Larochelle and M. Ranzato and R. Hadsell and M.F. Balcan and H. Lin},
  pages     = {4247--4258},
  publisher = {Curran Associates, Inc.},
  volume    = {33},
}

@InProceedings{ratenav2025,
  author    = {Li, Junjie and Zhang, Nan and Qu, Xiaoyang and Lu, Kai and Li, Guokuan and Wan, Jiguang and Wang, Jianzong},
  booktitle = {Findings of the {A}ssociation for {C}omputational {L}inguistics: {ACL} 2025},
  title     = {{RATE}-Nav: Region-Aware Termination Enhancement for Zero-shot Object Navigation with Vision-Language Models},
  year      = {2025},
  address   = {Vienna, Austria},
  editor    = {Che, Wanxiang and Nabende, Joyce and Shutova, Ekaterina and Pilehvar, Mohammad Taher},
  month     = jul,
  pages     = {6564--6574},
  publisher = {Association for Computational Linguistics},
  doi       = {10.18653/v1/2025.findings-acl.341},
  isbn      = {979-8-89176-256-5},
}

@InProceedings{cognav2025,
  author    = {Cao, Yihan and Zhang, Jiazhao and Yu, Zhinan and Liu, Shuzhen and Qin, Zheng and Zou, Qin and Du, Bo and Xu, Kai},
  booktitle = {Proceedings of the IEEE/CVF International Conference on Computer Vision (ICCV)},
  title     = {CogNav: Cognitive Process Modeling for Object Goal Navigation with LLMs},
  year      = {2025},
  month     = {October},
  pages     = {9550-9560},
}

@InProceedings{nie2025wmnav,
  author    = {Nie, Dujun and Guo, Xianda and Duan, Yiqun and Zhang, Ruijun and Chen, Long},
  booktitle = {2025 IEEE/RSJ International Conference on Intelligent Robots and Systems (IROS)},
  title     = {WMNav: Integrating Vision-Language Models into World Models for Object Goal Navigation},
  year      = {2025},
  pages     = {2392-2399},
  doi       = {10.1109/IROS60139.2025.11246684},
}

@Misc{anderson2018spl,
  author        = {Peter Anderson and Angel Chang and Devendra Singh Chaplot and Alexey Dosovitskiy and Saurabh Gupta and Vladlen Koltun and Jana Kosecka and Jitendra Malik and Roozbeh Mottaghi and Manolis Savva and Amir R. Zamir},
  title         = {On Evaluation of Embodied Navigation Agents},
  year          = {2018},
  archiveprefix = {arXiv},
  eprint        = {1807.06757},
  primaryclass  = {cs.AI},
  url           = {https://arxiv.org/abs/1807.06757},
}

@Misc{yokoyama2023esc,
  author        = {Kaiwen Zhou and Kaizhi Zheng and Connor Pryor and Yilin Shen and Hongxia Jin and Lise Getoor and Xin Eric Wang},
  title         = {ESC: Exploration with Soft Commonsense Constraints for Zero-shot Object Navigation},
  year          = {2023},
  archiveprefix = {arXiv},
  eprint        = {2301.13166},
  primaryclass  = {cs.AI},
  url           = {https://arxiv.org/abs/2301.13166},
}

@Misc{SOM2023,
  author        = {Jianwei Yang and Hao Zhang and Feng Li and Xueyan Zou and Chunyuan Li and Jianfeng Gao},
  title         = {Set-of-Mark Prompting Unleashes Extraordinary Visual Grounding in GPT-4V},
  year          = {2023},
  archiveprefix = {arXiv},
  eprint        = {2310.11441},
  primaryclass  = {cs.CV},
  url           = {https://arxiv.org/abs/2310.11441},
}

@InProceedings{yadav2023ovrl,
  author    = {Yadav, Karmesh and Ramrakhya, Ram and Ramakrishnan, Santhosh Kumar and Gervet, Theo and Turner, John and Gokaslan, Aaron and Maestre, Noah and Chang, Angel Xuan and Batra, Dhruv and Savva, Manolis and Clegg, Alexander William and Chaplot, Devendra Singh},
  booktitle = {Proceedings of the IEEE/CVF Conference on Computer Vision and Pattern Recognition (CVPR)},
  title     = {Habitat-Matterport 3D Semantics Dataset},
  year      = {2023},
  month     = {June},
  pages     = {4927-4936},
}

@Article{Douze2024,
  author   = {Douze, Matthijs and Guzhva, Alexandr and Deng, Chengqi and Johnson, Jeff and Szilvasy, Gergely and Mazaré, Pierre-Emmanuel and Lomeli, Maria and Hosseini, Lucas and Jégou, Hervé},
  journal  = {IEEE Transactions on Big Data},
  title    = {The Faiss Library},
  year     = {2026},
  number   = {2},
  pages    = {346-361},
  volume   = {12},
  doi      = {10.1109/TBDATA.2025.3618474},
}
}

\end{document}